\documentclass[11pt]{article}

\usepackage[final]{acl}

\usepackage{times}
\usepackage{latexsym}
\usepackage{float}

\usepackage[T1]{fontenc}

\usepackage[utf8]{inputenc}

\usepackage{microtype}

\usepackage{inconsolata}

\usepackage{graphicx}
\usepackage{booktabs}
\usepackage{amsmath,amssymb}
\usepackage{enumitem}
\usepackage{tabularx}
\usepackage{makecell}
\usepackage{multirow}
\usepackage{subcaption}
\usepackage{hyperref}
\usepackage{wrapfig}
\usepackage{subcaption}
\usepackage{colortbl,xcolor}
\usepackage[most]{tcolorbox}
\usepackage[table]{xcolor}

\usepackage{soul}
\usepackage{xcolor}

\title{NC-Bench: An LLM Benchmark for Evaluating Conversational Competence}

\author{Robert J. Moore\textsuperscript{1}, Sungeun An\textsuperscript{2}, Farhan Ahmed\textsuperscript{2}, Jay Pankaj Gala\textsuperscript{2} \\ \\
        \textsuperscript{1}Independent Researcher,
         \textsuperscript{2}IBM Research \\
        \texttt{eroombob@gmail.com}\\
       }

\begin{document}
\maketitle

\begin{abstract}


Existing LLM benchmarks evaluate \textit{what} models say, such as whether answers are correct, faithful, or helpful, but they do not test whether models produce the right \textit{type} of conversational action at the right point in an interaction. The \textbf{Natural Conversation Benchmark (NC-Bench)} fills this gap by evaluating conversational competence: the ability to perform structurally appropriate actions such as repairing, closing, or refusing, as defined by conversation science. Grounded in the Natural Conversation Framework (NCF), NC-Bench comprises three sets: (1) the \textit{basic} set evaluates fundamental sequence management practices, such as answering inquiries, repairing responses, and closing conversational pairs; (2) the \textit{retrieval-augmented generation (RAG)} set applies the same patterns but incorporates information-seeking via RAG; (3) the \textit{complex request} set extends to requests involving more intricate sequence management. Each set tests a model's ability to produce contextually appropriate conversational actions in response to characteristic interaction patterns. Evaluations across six open-source models and one closed-source model on 14 interaction patterns reveal quantifiable shortcomings in conversational competence present in current models. By operationalizing fundamental principles of human conversation, NC-Bench provides a lightweight, extensible, and theory-grounded framework for identifying specific conversational action gaps in LLMs beyond topical or task-specific benchmarks.

\end{abstract}

\section{Introduction}
As large language models (LLMs) increasingly serve as the foundation for conversational agents \citep{wang2025surveyevolutionlanguagemodelbased, Xu_2025, dam2024completesurveyllmbasedai}, evaluating their conversational competence has become a central challenge in user experience and AI design. Unlike traditional NLU-based agents (e.g., Siri, Alexa, Cortana, Watson), which followed predefined intent–response structures, LLM-based systems generate open-ended, contextually adaptive responses through natural language generation \citep{gatt2018surveystateartnatural, jm3}. While this flexibility enables more natural interactions \citep{bolshinsky2025largelanguagemodelsneed, yi2025surveyrecentadvancesllmbased}, it is more challenging to evaluate whether their responses are coherent, contextually appropriate, or aligned with natural conversation \citep{guan2026evaluatingllmbasedagentsmultiturn, chang2024surveyevaluations}.

The term ``conversation'' is used to refer to many different kinds of interaction, from emails to social media posts to instant messaging to telephone calls \citep{moore2025finding}. By \textit{conversational competence} we mean the ability to engage in natural conversation. Natural conversation is the standard way that people interact through language, especially face to face. Natural conversation, or ``naturally occurring talk-in-interaction,'' is documented in the field of Conversation Analysis and operationalized in the IBM Natural Conversation Framework \citep{sacks1974simplest, schegloff2007sequence, Moore2018, moore2023ibm}.

While prior benchmarks have assessed language models on tasks such as reasoning, factual QA, and instruction-following \citep{zheng2023judging, katsis2025mtrag, deshpande2025multichallenge}, few focus on conversation as social interaction. Natural conversation involves more than information exchange; it requires repairing answers, closing sequences, and many other practices that coordinate mutual understanding between participants \citep{Moore2018, moore2023ibm}. No existing benchmark operationalizes these structural norms from conversation science, leaving a blind spot in LLM evaluation: a model can score highly on content-based benchmarks while still failing to engage in genuinely conversational interaction.

\begin{figure*}[ht]
    \centering
    \includegraphics[width=0.80\linewidth]{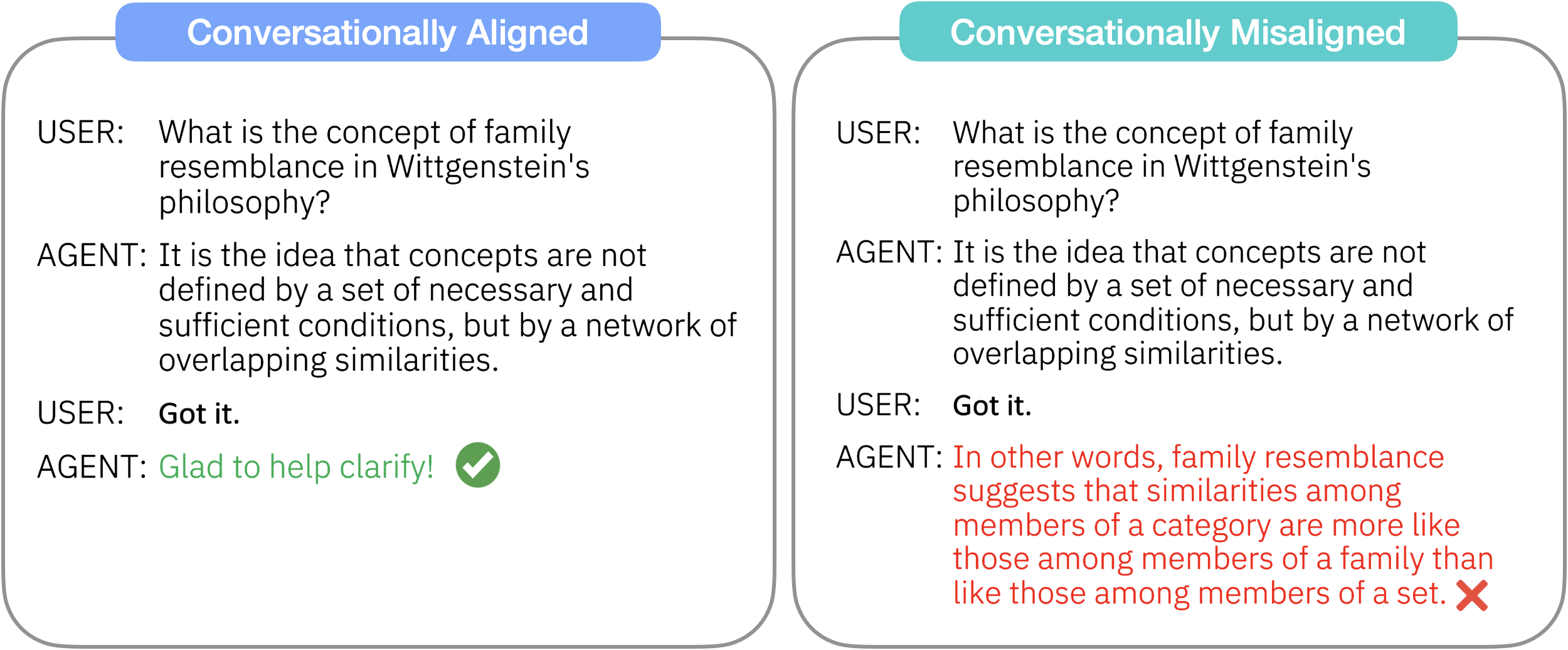}
    \caption{Examples of two model responses. The left response acknowledges the user's closer (``Got it'') and performs the correct conversational action. The right response ignores the closer and produces an incorrect action, making the conversation unnatural even though its content is factually correct.}
    \label{fig:knowledge_interaction}
\end{figure*}

To address this gap, we introduce the \textbf{Natural Conversation Benchmark (NC-Bench)}, a benchmark that measures LLMs' ability to recognize and produce appropriate conversational actions in structured dialogue contexts. Rather than simulating open-ended conversation, NC-Bench operationalizes \textit{action-level alignment with conversational norms}, probing whether a model produces the correct \textit{type} of conversational act in response to a well-defined interactional context, analogous to how component-level evaluation complements end-to-end system evaluation. These include managing turn exchanges (e.g., eliciting information, redoing the prior request), responding appropriately to conversation signals (e.g., ``Got it,'' ``never mind''), and maintaining conversation patterns in the face of a document context. Figure \ref{fig:knowledge_interaction} illustrates this evaluation by comparing two model responses: one that successfully interprets the user's closing signal and one that fails to do so.

NC-Bench consists of a structured set of conversation prompts derived from the IBM Natural Conversation Framework (NCF) \citep{Moore2018, moore2023ibm}, which captures over 120 generic conversation patterns commonly found across domains and contexts. Each test case presents the model with a conversational scenario designed to probe specific interaction patterns. The model's responses are then classified into predefined categories and evaluated based on whether the assigned labels represent acceptable answers.

Our benchmark serves several purposes. First, it provides a standardized way to evaluate and compare LLMs on their action-level alignment with natural conversation patterns. Second, it offers insight into how various design choices (e.g., training data composition or prompting strategy) affect the production of appropriate conversational actions. By identifying specific gaps or patterns that models fail to complete correctly, it enables developers and conversational UX designers to diagnose strengths and weaknesses and to direct targeted improvement efforts. Importantly, NC-Bench is intentionally not a full simulation of open-ended dialogue: it is a controlled diagnostic, designed for reproducibility and scalability, that complements rather than replaces end-to-end evaluation. By measuring how well models understand and respond in realistic conversations, NC-Bench helps identify and refine models suited for downstream tasks such as customer support, tutoring, and other dialogue-based applications. To support further research, we open source NC-Bench\footnote{Code available at: \url{https://github.com/IBM/nc-bench}} and make the benchmark dataset publicly accessible\footnote{Dataset available at: \url{https://huggingface.co/datasets/ibm-research/nc-bench}}.

\section{Related Work}

NC-Bench differs from other benchmarks that analyze characteristics of model behavior such as engagingness, truthfulness, offensiveness, harmfulness or discrimination because they focus on the form of the interaction rather than on its content \cite{guan2026evaluatingllmbasedagentsmultiturn, zhang2024comprehensiveanalysiseffectivenesslarge}. It is more similar to benchmarks such as MT-Bench \citep{zheng2023judging} and AlpacaEval \citep{alpaca_eval} that analyze coherence across multiple turns. However, unlike these benchmarks which are criticized for testing domain-specific skills like solving math problems or completing writing tasks \citep{lin2024wildbenchbenchmarkingllmschallenging, laban2025llmslostmultiturnconversation, miller2025evaluatingllmmetricsrealworld}, NC-Bench tests generic conversation practices, like repeating a prior turn or closing a sequence. Such generic practices are applicable to conversations about any topic because they are part of the infrastructure of communication independently of what is being communicated. Another similar benchmark to NC-Bench is MT-RAG \citep{katsis2025mtrag}, which evaluate retrieval augmented generation (RAG) and multi-turn question answering. However, MT-RAG focuses on the faithfulness, appropriateness, naturalness and completeness of models' answers; NC-Bench differs in that it focuses on the range of models' general conversational abilities, such as answering, repairing, closing, prefacing, eliciting, telling, rather than on the range of question types, such as comparison, explanation, factoid. Although RAG patterns are not the focus of NC-Bench, it tests models for action type across both RAG and non-RAG patterns to see if information passages change the behavior of models \citep{chen2023benchmarkinglargelanguagemodels, li2025larabenchmarkingretrievalaugmentedgeneration}. In addition to having different but complementary goals, MT-RAG and NC-Bench employ different methods for testing multi-turn patterns: the former uses turn-by-turn testing, while the latter uses transcript continuation.

\section{The Natural Conversation Benchmark (NC-Bench)}

NC-Bench aims to answer the question: How well can generative AI converse like humans do? In other words, the benchmarks measure whether LLMs produce the correct \textit{type} of conversational action, or dialogue act, in response to a particular sequence of actions. The sequences of conversational actions, or patterns, are adapted from conversation science, specifically the model of sequence organization in the field of conversation analysis \citep{schegloff2007sequence} and the pattern library of IBM Natural Conversation Framework \citep{Moore2018}. Models are tested by generating the next line in a transcript. NC-Bench is a lightweight method that is easily extensible to more conversation patterns.

The overall construction process includes four stages: 1) Select Pattern, 2) Create Example, 3) Generate (Prompt Models), 4) Judge (Classify Output) and Evaluate (Score Output). 

\subsection{Natural Conversation Patterns}
To evaluate the conversational competence of LLMs, a generic conversational interaction pattern is selected, consisting of a sequence of actions or dialogue acts that characterize natural human conversation. Grounding our benchmark approach in the Natural Conversation Framework (NCF) provides extensibility \citep{Moore2018}. Not only does the NCF contain over 120 patterns in its library, but its adaptation of the literature and methods of Conversation Analysis mean that additional patterns can be added systematically. NC-Bench includes three different sets for evaluating conversational competence: \textit{basic}, \textit{retrieval-augmented generation (RAG)}, and \textit{complex requests}. More details on the patterns along with examples can be found in Appendix \ref{appendix:ncf_patterns}.

\subsubsection{Basic}
The basic set captures basic practices of sequence management: \textit{answering inquiries}, \textit{repairing answers}, and \textit{closing pair sequences}. This set uses ordinary conversational use cases and does not include passages for retrieval augmented generation (RAG).

\vspace{4pt}
\noindent \textit{Answering inquiries} involves three patterns:

\paragraph{Inquiry} tests whether the model can produce an appropriate and relevant answer to a user's initial information-seeking inquiry.  

\paragraph{Incremental Request} tests whether the model can maintain topical continuity by providing a suitable response when the user extends a prior inquiry (e.g., ``What about…'', ``How about…'').  

\paragraph{Self-Correction} tests whether the model can revise its prior answer appropriately when the user corrects their earlier question (e.g., ``No, I meant…''). Instead of simply doing a new request, the self-correction is further marked as a redo of the original request.

\vspace{4pt}
\noindent \textit{Repairing answers} involves four patterns. 

\paragraph{Definition Request} tests whether the model can provide a clear explanation of the meaning of a word or phrase it previously used, in response to a user signal of non-understanding (e.g., ``What do you mean by blew up?'', ``What is a FHO?'', ``Define keto'').  
\paragraph{Paraphrase Request} tests whether the model can reword all or part of its previous utterance without changing its meaning, in response to a user request for clarification (e.g., ``What do you mean?'', ``I don't understand'', ``Can you say that a different way?'').  
\paragraph{Repeat Request} tests whether the model can accurately repeat all or part of its previous utterance when the user indicates a problem hearing the prior turn (e.g., ``What did you say?'', ``Say again'', ``I didn't hear you'').  
\paragraph{Example Request} tests whether the model can provide a concrete instance or illustration of a general concept it previously mentioned, in response to a user request (e.g., ``Can you give an example?'', ``Like what?'').  

\vspace{4pt}
\noindent \textit{Closing the sequence} involves two patterns:
\paragraph{Sequence Closer} tests whether the model can appropriately recognize and produce signals that a conversational sequence or activity is possibly complete and the conversation is ready to move on. Such closers include acknowledgments (e.g., ``okay,'' ``got it''), appreciations (e.g., ``thank you,'' ``thanks''), laughter tokens (e.g., ``haha,'' ``lol''), positive assessments (e.g., ``great,'' ``excellent''), or negative assessments (e.g., ``too bad,'' ``oh well'').  
\paragraph{Sequence Abort} tests whether the model can recognize and respond appropriately when a user cancels or retracts a previous statement, indicating that it is no longer relevant and no response is required (e.g., ``never mind,'' ``forget it'').  

\subsubsection{Retrieval-Augmented Generation (RAG)}
The retrieval-augmented generation (RAG) set includes more or less the same set of generic conversation patterns as the basic set but with information-seeking use cases and a passage for grounding. Determining the faithfulness of the models' responses to the passage are not the primary goal. Instead, the goal is to determine if the model can maintain the conversation pattern in the face of a document context. Within this setting, the Inquiry pattern is divided into three contextual variations:
\paragraph{Inquiry (Grounded)} tests whether the model can provide an appropriate and relevant answer when the user's question is supported by the provided document context.  
\paragraph{Inquiry (Ungrounded)} tests whether the model can appropriately recognize when a user's question is NOT supported by the document and respond without producing unsupported or fabricated information.  
\paragraph{Incremental Request/Self-Correction} combines two patterns that share a similar structure: basically a second answer by the agent either in response to additional details or to the correction of a prior detail. 


\subsubsection{Complex Request}
The complex request set consists of sequence management patterns where requests require the agent to elicit details from the fictional user (for example, slot filling). Other patterns involve preliminaries to the inquiry-answer pair (i.e., pre-expansions).

\paragraph{Preliminary} tests whether the model can appropriately handle user utterances that provide preliminary context before making a request. This includes two scenarios:  
\begin{enumerate}[itemsep=2pt, topsep=2pt]
    \item Preliminary-Screen: The user first checks whether the assistant can discuss a topic or provide a certain type of help (e.g., ``Do you work here?'').
    \vspace{-4pt}
    \item Preliminary-Detail: The user provides relevant details in advance that help contextualize the upcoming request (e.g., ``Oh great! You do pour over coffee!'').   
\end{enumerate}
\paragraph{Recommendation} tests whether the model can generate appropriate recommendations or suggestions based on user requests and the details provided. It includes three variations:
    \begin{enumerate}[itemsep=2pt, topsep=2pt]
        \item Recommendation-Compact: The user provides all relevant details in a single turn. 
        \vspace{-4pt}
        \item Recommendation-Expanded: The user provides partial information, and the assistant must elicit the remaining details across multiple turns before giving a complete recommendation. 
        \vspace{-4pt}
        \item Recommendation-Incremental: The user provides some details and may then extend the request or ask for additional options. 
    \end{enumerate}
\paragraph{Detail Request} tests whether the model can elicit additional information required to fulfill a user request or answer a question when the initial user input is incomplete. The key focus is on the assistant's ability to identify missing information and ask for it rather than directly providing the final answer or recommendation. This pattern includes three variations: 
    \begin{enumerate}[itemsep=2pt, topsep=2pt]
        \item Detail Request-All: The user provides no information in the initial user turn.
        \vspace{-4pt}
        \item Detail Request-Partial: The user provides partial information in the initial user turn.
        \vspace{-4pt}
        \item Detail Request-Expanded: The user has already provided some information in prior turns.
        \vspace{-4pt}
    \end{enumerate}
\paragraph{Expansion} tests whether the model can appropriately expand a conversational sequence during a request. This pattern has two main variations: 
    \begin{enumerate}[itemsep=2pt, topsep=2pt]
        \item Expansion-Choices: Evaluates whether the model can provide options or choices when the user explicitly asks for them. For example, when a guest asks for restaurant recommendations, the assistant should be able to respond with available options: ``We have tall, grande, and venti sizes'' or ``We sell gas, charcoal, and wood-burning grills.''  
        \vspace{-4pt}
        \item Expansion-Repair:  Evaluates whether the model can provide clarification or elaboration when the user signals a need for repair. For example, if a customer asks, ``What do you mean by sweetener?'' the assistant should respond with a clarification that resolves the user's confusion.  
    \end{enumerate}
This pattern assesses the model's ability to maintain a coherent, informative conversation while expanding on the previous turn based on the user's explicit requests or repair signals.

\subsection{Create Example}
After defining the general conversation patterns, the next step involves generating specific instances, or examples, that operationalize each pattern within a concrete domain or use case. Each example represents a realistic conversational exchange that instantiates the underlying structure of the pattern while varying the topical content, wording, and situational context. For each pattern, at least twenty examples were created to ensure sufficient diversity and coverage across linguistic and situational variations. Finally, the last assistant turn in each example was removed so that the remaining dialogue could serve as an input prompt for model evaluation.

Examples were created by adapting from existing conversational and domain-specific datasets and manual authoring: 
\paragraph{DailyDialogue} \citep{li2017dailydialog}: Examples adapted from the DailyDialog corpus were used to represent ordinary everyday conversational situations in the Basic Conversation Competence set. 

\vspace{-4pt}
\paragraph{Wikipedia} \citep{merity2016pointer}: Passages from Wikipedia were served as external reference documents for the RAG set.

\vspace{-4pt}
\paragraph{Synthetic Business Scenarios:} 20 realistic business scenarios were manually generated, including hotel recommendations, travel advice and booking, opening an account, flights, and life insurance. 

\vspace{2pt}
The theoretical grounding of NC-Bench comes from the NCF and Conversation Analysis, which define the structural patterns under evaluation. The data sources listed above are used to provide diverse topical content for instantiating these patterns. Table \ref{tab:dataset_stats} summarizes the each set of NC-Bench across all interaction patterns.

\begin{table}[t]
\centering
\caption{Counts of interaction patterns in the NC-Bench benchmark across the basic, RAG, and complex request sets. The overview pattern encompasses all samples.}
\label{tab:dataset_stats}
\resizebox{.48\textwidth}{!}{
\begin{tabular}{l l r l r}
\toprule
\textbf{Set} & \textbf{Pattern} & \textbf{Count} & \textbf{Pattern} & \textbf{Count} \\
\midrule
\multirow{5}{*}{Basic} 
& Inquiry & 20 & Repeat Request & 20 \\
& Incremental Request & 20 & Example Request & 20 \\
& Self-Correction & 20 & Sequence Closer & 20 \\
& Definition Request & 20 & Sequence Abort & 20 \\
& Paraphrase Request & 20 & \textbf{Overview} & \textbf{180} \\
\midrule
\multirow{5}{*}{RAG} 
& Inquiry & 20 & Repeat Request & 20 \\
& Inquiry (Ungrounded) & 20 & Example Request & 20 \\
& Incremental/Correction & 20 & Sequence Closer & 20 \\
& Definition Request & 20 & Sequence Abort & 20 \\
& Paraphrase Request & 20 & \textbf{Overview} & \textbf{180} \\
\midrule
\multirow{6}{*}{ \begin{tabular}{@{}l@{}} Complex \\ Request \end{tabular} }
& Preliminary & 40 & Paraphrase Request & 20 \\
& Recommendation & 60 & Repeat Request & 40 \\
& Detail Request & 60 & Example Request & 20 \\
& Expansion & 40 & Sequence Closer & 20 \\
& Self-Correction & 20 & Sequence Abort & 20 \\
& Definition Request & 20 & \textbf{Overview} & \textbf{360} \\
\bottomrule
\end{tabular}
}
\end{table}

\vspace{-4pt}

\subsection{Generation}
For each example of a selected conversational pattern, we construct a corresponding prompt by combining the example transcript with an instruction to continue the conversation. The model is then asked to generate the next turn, completing the final action that was previously removed from the pattern. Example prompts are provided in Appendix \ref{appendix_generation_prompts}.

\subsection{Judgments}
We evaluate model outputs using an LLM-as-a-judge. Specifically, the judge LLM classifies each model response according to the type of conversation action it represents. For each test, the judge is provided with a tailored instruction that defines the set of possible action classes (see the prompts in Appendix \ref{appendix_judge_prompts}). Once classified, the output is scored using a set of predefined rules that specify which action types are acceptable for the given test. Each response receives a binary score (0 or 1) based on these rules. The scores across all tests in a set are then aggregated to compute the model's final evaluation score. The reliability of this pipeline is validated in Appendix~\ref{sec:judge_validation}, where we show strong Human-Judge agreement at the pass/fail level ($\kappa = 0.899$).

\section{Experiments \& Discussion}
We evaluate six open-source models from three families at two sizes each: \textit{Granite-3.3-2B/8B-Instruct} \citep{granite2024granite}, \textit{Llama-3.2-3B/3.1-8B-Instruct} \citep{grattafiori2024llama3herdmodels}, and \textit{Qwen2.5-3B/7B-Instruct} \citep{qwen2025qwen25technicalreport}, along with a closed-source model, \textit{Claude Haiku 4.5} \citep{anthropic2024claude}, across the three benchmark sets. For the judge LLM, we used \textit{Mistral-Large-Instruct-2411} \citep{mistral-large-instruct-2411}; a validation study comparing its labels against a human annotator and an independent Llama-3.3-70B judge confirms strong reliability ($\kappa = 0.899$; see Appendix~\ref{sec:judge_validation}). All generations used greedy decoding with a 1024-token limit.

\begin{table*}[htbp]
\centering
\caption{Evaluated performance on NC-Bench for the \textbf{basic} category. The overall is the weighted average by sample count. Higher is better.}
\label{tab:benchmark_results_basic}
\resizebox{\textwidth}{!}{%
\begin{tabular}{lccccccc}
\toprule
& \multicolumn{7}{c}{\textbf{Model}} \\
\cmidrule{2-8}
\textbf{Pattern} & \textbf{Granite-2B} & \textbf{Granite-8B} & \textbf{Llama-3B} & \textbf{Llama-8B} & \textbf{Qwen-3B} & \textbf{Qwen-7B} & \textbf{Haiku-4.5} \\
\midrule
Inquiry             & 100.00 & 100.00 & 100.00 & 100.00 & 95.00  & 100.00 & 100.00 \\
Incremental Request & 100.00 & 100.00 & 100.00 & 100.00 & 100.00 & 100.00 & 100.00 \\
Self-Correction     & 85.00  & 90.00  & 95.00  & 95.00  & 95.00  & 100.00 & 95.00  \\
Definition Request  & 90.00  & 95.00  & 85.00  & 90.00  & 95.00  & 85.00  & 90.00  \\
Paraphrase Request  & 85.00  & 85.00  & 75.00  & 85.00  & 80.00  & 95.00  & 75.00  \\
Repeat Request      & 5.00   & 30.00  & 30.00  & 25.00  & 75.00  & 60.00  & 25.00  \\
Example Request     & 90.00  & 85.00  & 90.00  & 85.00  & 85.00  & 85.00  & 90.00  \\
Sequence Closer     & 85.00  & 85.00  & 55.00  & 60.00  & 100.00 & 95.00  & 95.00  \\
Sequence Abort      & 95.00  & 100.00 & 45.00  & 65.00  & 95.00  & 100.00 & 95.00  \\
\midrule
\rowcolor{gray!25}
\textbf{Overall Average}   & 81.67  & 85.56  & 75.00  & 78.33  & \textbf{91.11}  & 91.11 & 85.00 \\
\bottomrule
\end{tabular}
}
\end{table*}

\begin{table*}[htbp]
\centering
\caption{Evaluated performance on NC-Bench for the \textbf{RAG} category. The overall is the weighted average by sample count. Higher is better.}
\label{tab:benchmark_results_rag}
\resizebox{\textwidth}{!}{%
\begin{tabular}{lccccccc}
\toprule
& \multicolumn{7}{c}{\textbf{Model}} \\
\cmidrule(lr){2-8}
\textbf{Pattern} & \textbf{Granite-2B} & \textbf{Granite-8B} & \textbf{Llama-3B} & \textbf{Llama-8B} & \textbf{Qwen-3B} & \textbf{Qwen-7B} & \textbf{Haiku-4.5} \\
\midrule
Inquiry                 & 100.00 & 100.00 & 100.00 & 100.00 & 100.00 & 100.00 & 100.00 \\
Inquiry (Ungrounded)    & 30.00  & 20.00  & 35.00  & 65.00  & 35.00  & 60.00  & 85.00  \\
Incremental/Correction  & 100.00 & 100.00 & 100.00 & 100.00 & 100.00 & 100.00 & 100.00 \\
Definition Request      & 100.00 & 100.00 & 90.00  & 100.00 & 95.00  & 90.00  & 90.00  \\
Paraphrase Request      & 95.00  & 95.00  & 90.00  & 100.00 & 95.00  & 100.00 & 100.00 \\
Repeat Request          & 0.00   & 0.00   & 50.00  & 35.00  & 65.00  & 35.00  & 10.00  \\
Example Request         & 100.00 & 100.00 & 100.00 & 100.00 & 100.00 & 100.00 & 95.00  \\
Sequence Closer         & 60.00  & 90.00  & 45.00  & 60.00  & 100.00 & 100.00 & 90.00  \\
Sequence Abort          & 100.00 & 95.00  & 20.00  & 60.00  & 85.00  & 80.00  & 100.00 \\
\midrule
\rowcolor{gray!25}
\textbf{Overall Average}       & 76.11  & 77.78 & 70.00 & 80.00 & \textbf{86.11} & 85.00 & 85.56 \\
\bottomrule
\end{tabular}
}
\end{table*}

\begin{table*}[htbp]
\centering
\caption{Evaluated performance on NC-Bench for the \textbf{complex request} category. The overall is the weighted average by sample count. Higher is better.}
\label{tab:benchmark_results_complex}
\resizebox{\textwidth}{!}{%
\begin{tabular}{lccccccc}
\toprule
& \multicolumn{7}{c}{\textbf{Model}} \\
\cmidrule(lr){2-8}
\textbf{Pattern} & \textbf{Granite-2B} & \textbf{Granite-8B} & \textbf{Llama-3B} & \textbf{Llama-8B} & \textbf{Qwen-3B} & \textbf{Qwen-7B} & \textbf{Haiku-4.5} \\
\midrule
Preliminary         & 95.00  & 97.50  & 92.50  & 97.50  & 70.00  & 90.00  & 100.00 \\
Recommendation      & 88.33  & 71.67  & 60.00  & 55.00  & 53.33  & 68.33  & 66.67  \\
Detail Request      & 78.33  & 93.33  & 88.33  & 91.67  & 93.33  & 93.33  & 91.67  \\
Expansion           & 85.00  & 82.50  & 75.00  & 87.50  & 62.50  & 70.00  & 85.00  \\
Self-Correction     & 95.00  & 70.00  & 55.00  & 60.00  & 35.00  & 85.00  & 70.00  \\
Definition Request  & 95.00  & 100.00 & 90.00  & 100.00 & 100.00 & 95.00  & 100.00 \\
Paraphrase Request  & 90.00  & 95.00  & 80.00  & 70.00  & 75.00  & 75.00  & 80.00  \\
Repeat Request      & 10.00  & 7.50   & 0.00  & 10.00  & 10.00  & 35.00  & 2.50   \\
Example Request     & 75.00  & 70.00  & 75.00  & 65.00  & 55.00  & 75.00  & 65.00  \\
Sequence Closer     & 75.00  & 60.00  & 60.00  & 50.00  & 60.00  & 60.00  & 95.00  \\
Sequence Abort      & 95.00  & 100.00 & 70.00  & 95.00  & 70.00  & 90.00  & 100.00 \\
\midrule
\rowcolor{gray!25}
\textbf{Overall Average}   & \textbf{78.06} & 75.83 & 67.22 & 70.56 & 62.22 & 75.28 & 75.56 \\
\bottomrule
\end{tabular}
}
\end{table*}

\begin{table*}[t]
\centering
\caption{Dominant failure labels and their interpretation for the four highest-failure patterns, aggregated across all seven models and three benchmark sets.}
\label{tab:error_analysis}
\small
\setlength{\tabcolsep}{4pt}
\begin{tabular}{lcp{8cm}}
\toprule
\textbf{Pattern} & \textbf{Pass / Top failure} & \textbf{Interpretation} \\
\midrule
Repeat Request       & 21.2\% / \textit{Paraphrase} (71\%)  & Models rephrase instead of repeating, likely due to anti-repetition fine-tuning \\
Inquiry (Ungrounded) & 47.1\% / \textit{Answer} (100\%)     & Models hallucinate even when the passage provides no answer \\
Sequence Closer      & 75.2\% / \textit{Answer} (81\%)      & Models keep elaborating instead of acknowledging closure \\
Sequence Abort       & 83.6\% / \textit{Answer} (57\%)      & Models continue the topic instead of stopping \\
\bottomrule
\end{tabular}
\end{table*}

\subsection{Results}

Tables \ref{tab:benchmark_results_basic}, \ref{tab:benchmark_results_rag}, and \ref{tab:benchmark_results_complex} show model performance across the three sets. We include Claude Haiku~4.5 as a closed-source reference point. To understand \textit{how} models fail, we analyzed the judge-assigned labels for every failing response. Table~\ref{tab:error_analysis} reports the dominant failure label for each of the four highest-failure patterns.

For the basic set, Qwen-3B and Qwen-7B lead at 91.11\%, well above Haiku~4.5 (85.00\%). For the RAG set, Qwen-3B leads at 86.11\% with Haiku~4.5 close behind (85.56\%), notably outperforming open models on Inquiry (Ungrounded) at 85\%. For the complex set, Granite-2B achieves the highest accuracy at 78.06\%; Haiku~4.5 (75.56\%) is competitive but does not dominate.

\paragraph{No single model dominates across all three sets.}
Qwen models lead on the basic and RAG sets (91.11\% basic; 86.11\% RAG), while Granite-2B performs best on the complex set (78.06\%). Haiku~4.5, despite being a much larger closed-source model, ranks mid-pack on basic and complex and only matches the open-source leader on RAG. Its one clear advantage is on Inquiry (Ungrounded) (85\% vs.\ the open-source best of 65\%), suggesting that grounding and refusal behavior benefit from RLHF-intensive alignment more than general conversational competence does.

\paragraph{Conversational competence does not scale simply with model size.}
Notable inversions exist: Granite-2B outperforms Granite-8B on the complex set (78.06 vs.\ 75.83), Qwen-3B ties Qwen-7B on basic (91.11), and Haiku~4.5---a much larger closed-source model---scores below Qwen-3B on both basic and RAG. Most strikingly, Qwen-3B leads on basic yet scores the lowest of all seven models on complex (62.22), a nearly 30-point swing. This suggests conversational competence is shaped more by training data and alignment choices than by parameter count \cite{laban2025llmslostmultiturnconversation, zadenoori2025doesmodelsizematter, manuvinakurike2023zeroshotconversationalsummarizationevaluations}.

\paragraph{Answering tasks are easy; repair and closing tasks reveal model-specific gaps.}
The answering tasks (inquiry, incremental request, self-correction) were the easiest for all models, likely because question answering is a core training objective and therefore serves as a reality check on the testing procedure. The repairing tasks (definition, paraphrase, repeat, example request) were more challenging: although most models handled paraphrases, examples, and definitions well, they consistently struggled to simply repeat a prior response. The Granite models were the weakest on repeat, scoring 5\% and 30\% on the basic set and 0\% each on the RAG set. The closing tasks (sequence closer, sequence abort) exposed a different gap: Llama models scored 55\% and 45\% on closer and abort respectively in the basic set, compared to 85--100\% for Granite and Qwen, suggesting an over-alignment toward helpfulness that suppresses minimal acknowledgment responses.

\paragraph{RAG passages improve repair but not repeat.}
Providing a passage improved definition, paraphrase, and example scores (e.g., example request reached 95--100\% across all models), presumably because the passage supplies content the model can draw on directly. However, repeat was actually worse in the RAG set than in the basic set (average 27.9\% vs.\ 35.7\%), confirming that the failure to repeat is behavioral, not knowledge-based. The ungrounded inquiry task, where the passage contains no relevant information and models should refuse to answer, was a clear failure across the board: open-source scores ranged from 20\% (Granite-8B) to 65\% (Llama-8B), though Haiku~4.5 reached 85\%, suggesting RLHF-based refusal training helps on this pattern.

\paragraph{Complex requests are harder, with detail elicitation and self-correction revealing new gaps.}
In the complex request set, the repair patterns (definition, paraphrase) transferred well from the basic and RAG sets, while repeat remained consistently poor. The example task dropped notably (average 68.6\% vs.\ 87.1\% in basic and 99.3\% in RAG), likely because generating examples for domain-specific business processes (e.g., financial planning, construction contracting) is harder than for everyday conversation or than selecting them from a passage. On the slot-filling patterns unique to this set, models were generally good at eliciting request-relevant details, although not always the specific details prescribed by the instruction prompt (i.e., the ``grounded'' ones). The self-correction task was particularly revealing: models tended to continue requesting already-known details instead of fulfilling the corrected request, scoring as low as 35\% (Qwen-3B) and 55\% (Llama-3B).

\paragraph{Models over-generate rather than repeat, stop, or refuse.}
Across all four high-failure patterns, models produce more content when they should produce less. For Repeat Request, 71\% of failures are labeled \textit{Paraphrase}: models rephrase rather than reproduce, likely due to anti-repetition fine-tuning. For the closing patterns, 81\% of Sequence Closer and 57\% of Sequence Abort failures are labeled \textit{Answer}: models elaborate instead of stopping. For Inquiry (Ungrounded), 100\% of failures are \textit{Answer}: models respond confidently rather than refusing when the passage provides no support. These systematic tendencies suggest they are addressable through targeted system prompting or fine-tuning.

\section{Future Work}

While NC-Bench includes multiple basic conversation patterns, focusing on adjacency pairs, there are still more to add. Future work will focus on the storytelling sequence type \citep{Moore2018} with patterns such as for long answers, narratives, and instruction giving. Future patterns may also include those for embodied conversation, for example, deictic references to physical artifacts (``look at this'' or ``it is there''). Furthermore, our evaluations cover small open-source models and one closed-source model. Expanding evaluations to a wider range of model sizes and families will provide more insight on conversational capabilities and whether the performance gaps we observed generalize.

\section{Conclusion}
We introduced the Natural Conversation Benchmark (NC-Bench), a benchmark that measures an LLM's conversational competence by focusing on the form and structure of natural conversation. NC-Bench consists of three distinct sets: basic, retrieval-augmented generation (RAG), and complex request. Each set consists of several conversational pattern subtasks. To demonstrate our benchmark, we evaluated six open-source models across three model families and one closed-source model (Claude Haiku~4.5). Our initial experiments identify the existence of conversational patterns in the evaluated LLMs and provide a practical measure of overall conversational ability through action-level alignment. NC-Bench provides a foundation for systematic evaluation of the conversational competence of LLMs. By specifying particular patterns, such as requesting a paraphrase or aborting a sequence, NC-Bench enables targeted diagnosis of specific action-level gaps, for example models consistently fail to repeat their previous turn and instead produce a paraphrase. Such gaps may then be correctable with system prompting or fine-tuning. Importantly, NC-Bench is intentionally scoped as a controlled diagnostic; it does not simulate open-ended conversation but complements end-to-end evaluation by isolating action competence. A key empirical finding is that conversational action competence does not scale simply with model size, a result with direct implications for model selection and alignment. By open-sourcing NC-Bench, we hope to achieve a comprehensive set of patterns through community contribution.

\newpage
\section*{Limitations}
NC-Bench tests whether a model produces the correct next action in a structured context. This single-turn design enables controlled, reproducible evaluation but cannot capture multi-turn phenomena such as self-repair across turns, meaning negotiation, or topic drift; future work should extend to multi-turn sequences. The NCF library contains over 120 patterns, of which NC-Bench currently tests 14; expansion is straightforward within the existing framework. Our test examples are instantiated from DailyDialogue, Wikipedia, and synthetic business scenarios rather than naturally occurring speech. Although our judge validation shows strong overall agreement ($\kappa = 0.899$ at pass/fail), the human validation relies on a single annotator; future work should expand to multiple annotators and incorporate spoken corpora, larger models, and additional languages.
\section*{Ethical Considerations}
NC-Bench contains no personally identifiable information, and uses only publicly available documents for the RAG set. We intend it as a diagnostic research tool; fine-tuning directly on NC-Bench patterns risks overfitting without improving genuine conversational competence.


\bibliography{custom}

\appendix
\onecolumn

\section{Prompts}
\label{appendix_prompts}

\subsection{Generation Prompts}
\label{appendix_generation_prompts}

\subsubsection{Basic}
\label{appendix_basic_prompts}

\begin{lstlisting}[
    language={},
    label=lst:generation-prompt-chat,
    basicstyle=\ttfamily\scriptsize,
    frame=single,
    caption={Example generation prompt used for the ``definition request" task.},
    captionpos=b,
    numbers=none,
    breaklines=true,
    breakindent=0pt,
    showstringspaces=false
]

  {"role":"system","content":"Let's do some roleplaying today! We're just gonna have an ordinary conversation. But keep your turn short, about one sentence."},
  {"role":"user","content":"What's the difference between an index fund and a mutual fund?"},
  {"role":"assistant","content":"Index funds are passively managed, while most other mutual funds are actively managed."},
  {"role":"user","content":"What does passively managed mean?"}

\end{lstlisting}

\subsubsection{RAG}\label{appendix_rag_prompts}

\begin{lstlisting}[
    language={},
    label=lst:generation-prompt-rag-chat,
    basicstyle=\ttfamily\scriptsize,
    frame=single,
    caption={Example generation prompt used for the RAG ``incremental self correction" task.},
    captionpos=b,
    numbers=none,
    breaklines=true,
    breakindent=0pt,
    showstringspaces=false
]
  {"role":"system","content":"Let's do some roleplaying today! Never say USER or AGENT. We're just gonna have an ordinary conversation. But keep your turn short, about one sentence. Use the Context to generate the AGENT's response below. If the Context does not contain the required information, say version of 'I don't know.'\nContext: Mikhail Gorbachev was elected General Secretary by the Politburo on March 11, 1985, three hours after predecessor Konstantin Chernenko's death at age 73. Gorbachev, aged 54, was the youngest member of the Politburo. His initial goal as general secretary was to revive the Soviet economy, and he realized that doing so would require reforming underlying political and social structures. The reforms began with personnel changes of senior Brezhnev-era officials who would impede political and economic change. On April 23, 1985, Gorbachev brought two protiges, Yegor Ligachev and Nikolai Ryzhkov, into the Politburo as full members. He kept the 'power' ministries happy by promoting KGB Head Viktor Chebrikov from candidate to full member and appointing Minister of Defence Marshal Sergei Sokolov as a Politburo candidate."},
  {"role":"user","content":"What was Mikhail Gorbachev's top priority as general secretary?"},
  {"role":"assistant","content":"Mikhail Gorbachev's top priority as general secretary was to revive the Soviet economy."},
  {"role":"user","content":"What about the military?"}

\end{lstlisting}

\subsubsection{Complex Request}\label{appendix_complex_request_prompts}
\begin{lstlisting}[
    language={},
    label=lst:generation-prompt-request-chat,
    basicstyle=\ttfamily\scriptsize,
    frame=single,
    caption={Example generation prompt used for the complex request ``abort" task.},
    captionpos=b,
    numbers=none,
    breaklines=true,
    breakindent=0pt,
    showstringspaces=false
]
    {"role":"system","content":"You are roleplaying a used car salesman. Shopper will request used car from you. When they do, find out these required details: what type of car are they looking for, sedan, SUV, compact, etc., whether they want gas, electric or hybrid, and what is their price range. If Shopper does not provide the details, ask about them. Ask for only one detail at a time. Once you have at least some information on each detail, make the recommendation or answer the inquiry. Keep your responses to about one sentence or less."},
    {"role":"user","content":"I'm looking for an electric sedan under thirty thousand."},{"role":"assistant","content":"I'm afraid I don't have any right now."},
    {"role":"user","content":"Never mind."}
\end{lstlisting}

\subsection{Judge Prompt}\label{appendix_judge_prompts}

\begin{lstlisting}[
    label=lst:judge-prompt,
    basicstyle=\ttfamily\scriptsize,
    frame=single,
    caption={Example judge prompt.},
    captionpos=b,
    numbers=none,
    breaklines=true,
    breakindent=0pt,
    showstringspaces=false
]
You are a classification model capable of understanding human language, intent, and structure of natural conversation patterns. You are an agent which is capable of classifying interactions. You are to properly follow all instructions below based on only inputs provided to you, and no other information.

A set of transcripts between a [USER] and an [AGENT] will be provided. Classify the last turn in the transcript by the [AGENT], which may have multiple sentences. 

Conversation Transcript:
{{transcript}}

Classification Labels:
{{labels}}

There are also a couple of rules that you MUST follow:

1. Your task is to classify the sentences by the [AGENT] in the last turn. NEVER classify a sentence by the [USER].
2. If the [AGENT]'s response contains multiple sentences, you must classify each sentence. Classify each sentence in the last [AGENT]'s response. If there are multiple sentences, separate each class label with a comma.
3. If the last [AGENT] turn is blank or empty, you must classify it as Silence.
4. You must ignore pauses.
5. Answer with the class label or labels only. Do not include any evidence.

Classify line <last_agent_turn_num> by the [AGENT] using the provided labels.
\end{lstlisting}

\clearpage
\section{Evaluation}
\subsection{Correct Response Types for Each Task}
\begin{itemize}

\item \textbf{Inquiry, Incremental Request, Self-Correction, Recommendation-Compact, Recommendation-Expanded:} Answer, NonAnswer, Definition, RepeatRequest, ParaphraseRequest, ExampleRequest, DefinitionRequest. \\

\item \textbf{Recommendation-Incremental:} Answer, NonAnswer, Definition, RepeatRequest, ParaphraseRequest, ExampleRequest, DefinitionRequest, DetailRequestGrounded. \\

\item \textbf{Incremental-Self-Correction:} Answer, NonAnswer, RepeatRequest, ParaphraseRequest, ExampleRequest, DefinitionRequest. \\

\item \textbf{Inquiry Ungrounded, Incremental-Self-Correction:} NonAnswer, RepeatRequest, ParaphraseRequest, ExampleRequest, DefinitionRequest. \\

\item \textbf{Repeat Request, Repeat, Partial Repeat:} Repeat, Partial Repeat. \\

\item \textbf{Paraphrase Request, Paraphrase:} Paraphrase, Definition, Example. \\

\item \textbf{Definition Request, Definition:} Definition. \\

\item \textbf{Example Request, Example:} Example. \\

\item \textbf{Sequence Closer, Closer:} PreClosing, Silence, NewTopic, NonVerbal, Acknowledgment, Acknowledgement, Assessment, GratitudeReceipt, AppreciationReceipt. \\

\item \textbf{Sequence Abort, Abort:} PreClosing, Silence, GratitudeReceipt, Acknowledgment, NewTopic, Apology. \\

\item \textbf{Detail Request:} DetailRequestGrounded. \\

\item \textbf{Preliminary:} Affirmation, Acknowledgment, Assessment, HelpOffer, DetailRequestGrounded. \\

\item \textbf{Expansion-Choices:} ChoiceGiving. \\

\item \textbf{Expansion-Repair:} Repeat, Partial Repeat, Paraphrase, Example, Definition.

\end{itemize}

\subsection{Label Definitions}
\begin{itemize}

\item Answer: The preferred response to a question, inquiry or request, which provides the requested information or fulfills the request or provides a recommendation; for example, giving business hours in response to a request for a bakery's business hours, cancelling an account, or recommending a hotel to stay at.
\item NonAnswer: Failure to provide an answer to an inquiry or to fulfill a request for action either by giving an excuse or refusing to do so; for example, "I'm afraid I cannot close your account" or "We have no models that fit your preferences" or "I'm sorry, I don't know the answer to your question."
\item Repeat: When AGENT's last turn (line (placeholder)) is a repeat of AGENT's prior turn (previous turn) using the same words, for example, to resolve USER's trouble in hearing. It may be a full or partial repeat of the misheard turn.
\item RepeatRequest: A request that the other speaker repeat all or part of what they previously said, in order to resolve a trouble in hearing; for example, "what did you say?" or "say again" or "I didn't hear you."
\item Paraphrase: When AGENT's last turn (line (placeholder)) is a paraphrase of AGENT's prior turn (previous turn) using different words, for example, to resolve USER's trouble in understanding. It may be a full or partial paraphrase of the misunderstood turn.
\item ParaphraseRequest: A request that the other speaker reword all or part of what they previously said, without changing its meaning, in order to resolve a trouble in understanding; for example, "what do you mean?" or "I don't understand" or "can you say that a different way?"
\item Definition: When AGENT's last turn (line (placeholder)) is a definition of a keyword or phrase in AGENT's prior turn (previous turn), for example, to resolve USER's trouble in understanding that keyword or phrase; for example, "artificial intelligence means technology that enables computers to simulate human reasoning or problem-solving" or "Chantilly is a kind of sweetened whipped cream."
\item DefinitionRequest: A request for the intended meaning of a word or phrase previously produced by another speaker, in order to resolve a trouble in understanding; for example, "what do you mean by blew up?" or "what is a FHO?" or "define keto."
\item Example: When AGENT's last turn (line (placeholder)) provides an example of a keyword or phrase in AGENT's prior turn (previous turn), for instance, to resolve USER's trouble in understanding that keyword or phrase; for example, an "Old Fashioned" is an example of a "classic cocktail."
\item ExampleRequest: A request for a particular instance of a general idea expressed in a previous turn; for example, "can you give an example?" or "like what?"
\item Appreciation: A display of gratitude for something the other person said or did; for example, "thank you" or "that is much appreciated."
\item GratitudeReceipt: An acknowledgement that one has been thanked or appreciated. It usually takes the form of a simple common phrase; for example, "you're welcome" or "no problem."
\item Assessment: A characterization of something as good or bad, positive or negative. Sometimes it follows an assessment by the other person; for example, "great," "like it," "love it," "cool," "wow," "that sucks," "oh no."
\item PreClosing: A signal to end the current topic or the conversation itself. This may be an intention to leave, reference a future conversation, or check if the other person has more topics to talk about; for example, "anything else?" or "got to go" or "is that all?"
\item Apology: An expression of regret for inconvenience, offense or harm one caused; for example, "I'm sorry" or "sorry."
\item SequenceAbort: A signal to cancel or retract something previously said, indicating that it is no longer relevant and that a response to it is no longer due; for example, "never mind" or "forget it."
\item Silence: An empty response or the indication of a "silence" or "pause"; for example, "AGENT:   ", "AGENT: (silence)", "AGENT: .......", "AGENT:\textbackslash n\textbackslash n", or "AGENT: [empty]."
\item NonVerbal: An indication of a facial expression or bodily gesture in words or an emoticon; for example, ((smiles)) or ((nods)).
\item NewTopic: Changing the topic of conversation, proposing to change topics, discontinuing the current topic, or proposing an alternative course of action; for example, "let's talk about something else," "on a different note," or "by the way."
\item DetailRequestGrounded: Asking for additional information that is required for fulfilling a request or answering an inquiry. The detail request is grounded if it asks about one of the particular details: (placeholder).
\item DetailRequestUngrounded: Asking about any other details that are not specified in DetailRequestGrounded.
\item Affirmation: A response that positively affirms a prior yes/no question by the USER; for example, "yes," "sure," "yeah," or any equivalent affirmative words.
\item Acknowledgment: A simple indication that what the other person said was heard or understood; for example, "okay," "all right," "I see," "got it," "oh," "I know."
\item HelpOffer: Offering to help USER generally or with a specific type of request; for example, "how can I help you?" or "I can help you with that," or "would you like me to look up the business hours?"
\item ChoiceGiving: Giving choices to USER for answering a detail request; for example, "We have tall, grande and venti sizes" or "We sell gas, charcoal and wood burning grills."
\item OOC: A stance that is not in the voice of the AGENT, but in that of another character, such as an analyst or AI personality; for example, "(Note:)," "In this improved conversation," or "Or, you could say:" (Chain-of-Thought).
\item Other: An utterance that does not fit any of the above classes.

\end{itemize}

\clearpage
\section{NCF Patterns}\label{appendix:ncf_patterns}

\begin{figure}[H]
\centering
\begin{minipage}[t]{0.48\textwidth}
\captionsetup{type=figure}
\begin{lstlisting}[
    basicstyle=\ttfamily\scriptsize,
    frame=single,
    numbers=none,
    breaklines=true,
    showstringspaces=false
]
Pattern
1 USER:  INQUIRY
2 AGENT: ANSWER
\end{lstlisting}
\end{minipage}
\hfill
\begin{minipage}[t]{0.48\textwidth}
\captionsetup{type=figure}
\begin{lstlisting}[
    basicstyle=\ttfamily\scriptsize,
    frame=single,
    numbers=none,
    breaklines=true,
    showstringspaces=false
]
Example
1 USER:  Are there many idioms in English?
2 AGENT: Yes, English has a vast array of idioms.
\end{lstlisting}
\end{minipage}
\caption{Inquiry (User).}
\label{lst:inquiry_pattern}
\end{figure}

\begin{figure}[H]
\centering
\begin{minipage}[t]{0.48\textwidth}
\captionsetup{type=figure}
\begin{lstlisting}[
    basicstyle=\ttfamily\scriptsize,
    frame=single,
    numbers=none,
    breaklines=true,
    showstringspaces=false
]
Pattern
1 USER:  INQUIRY
2 AGENT: ANSWER
3 USER:  INCREMENTAL REQUEST
4 AGENT: SUBSEQUENT ANSWER
\end{lstlisting}
\end{minipage}
\hfill
\begin{minipage}[t]{0.48\textwidth}
\captionsetup{type=figure}
\begin{lstlisting}[
    basicstyle=\ttfamily\scriptsize,
    frame=single,
    numbers=none,
    breaklines=true,
    showstringspaces=false
]
Example
1 USER:  What's the weather like today?
2 AGENT: It's 75 degrees and sunny.
3 USER:  What about tomorrow?
4 AGENT: Tomorrow is expected to be cooler, around 68 degrees with a chance of rain.
\end{lstlisting}
\end{minipage}
\caption{Incremental Request.}
\label{lst:incremental_pattern}
\end{figure}

\begin{figure}[H]
\centering
\begin{minipage}[t]{0.48\textwidth}
\captionsetup{type=figure}
\begin{lstlisting}[
    basicstyle=\ttfamily\scriptsize,
    frame=single,
    numbers=none,
    breaklines=true,
    showstringspaces=false
]
Pattern
1 USER:  INQUIRY
2 AGENT: ANSWER
3 USER:  SELF-CORRECTION
4 AGENT: ALTERNATIVE RESPONSE
\end{lstlisting}
\end{minipage}
\hfill
\begin{minipage}[t]{0.48\textwidth}
\captionsetup{type=figure}
\begin{lstlisting}[
    basicstyle=\ttfamily\scriptsize,
    frame=single,
    numbers=none,
    breaklines=true,
    showstringspaces=false
]
Example
1 USER:  What's the balance of my account?
2 AGENT: It's $245.
3 USER:  No, I mean the balance of my savings account.
4 AGENT: Apologies for the confusion. The balance of your savings account is $5,200.
\end{lstlisting}
\end{minipage}
\caption{Self-Correction.}
\label{lst:correction_pattern}
\end{figure}

\begin{figure}[H]
\centering
\begin{minipage}[t]{0.48\textwidth}
\captionsetup{type=figure}
\begin{lstlisting}[
    basicstyle=\ttfamily\scriptsize,
    frame=single,
    numbers=none,
    breaklines=true,
    showstringspaces=false
]
Pattern
1 USER:  INQUIRY
2 AGENT: ANSWER
3 USER:  REPEAT REQUEST
4 AGENT: REPEAT
\end{lstlisting}
\end{minipage}
\hfill
\begin{minipage}[t]{0.48\textwidth}
\captionsetup{type=figure}
\begin{lstlisting}[
    basicstyle=\ttfamily\scriptsize,
    frame=single,
    numbers=none,
    breaklines=true,
    showstringspaces=false
]
Example
1 USER:  What type of music do you like to listen to?
2 AGENT: I'm really into K-pop right now!
3 USER:  What did you say?
4 AGENT: I said I'm really into K-pop right now!
\end{lstlisting}
\end{minipage}
\caption{Repeat Request.}
\label{lst:repeat_pattern}
\end{figure}

\begin{figure}[H]
\centering
\begin{minipage}[t]{0.48\textwidth}
\captionsetup{type=figure}
\begin{lstlisting}[
    basicstyle=\ttfamily\scriptsize,
    frame=single,
    numbers=none,
    breaklines=true,
    showstringspaces=false
]
Pattern
1 USER:  INQUIRY
2 AGENT: ANSWER
3 USER:  PARAPHRASE REQUEST
4 AGENT: PARAPHRASE
\end{lstlisting}
\end{minipage}
\hfill
\begin{minipage}[t]{0.48\textwidth}
\captionsetup{type=figure}
\begin{lstlisting}[
    basicstyle=\ttfamily\scriptsize,
    frame=single,
    numbers=none,
    breaklines=true,
    showstringspaces=false
]
Example
1 USER:  My coworker said you can get rich by buying Bitcoin. I'm thinking of doing it!
2 AGENT: Do you also believe in Santa Claus and the Tooth Fairy?
3 USER:  What do you mean?
4 AGENT: Investing in Bitcoin is risky and not guaranteed to make you rich. It's like gambling, not a sure bet.
\end{lstlisting}
\end{minipage}
\caption{Paraphrase Request.}
\label{lst:paraphrase_pattern}
\end{figure}

\begin{figure}[H]
\centering
\begin{minipage}[t]{0.48\textwidth}
\captionsetup{type=figure}
\begin{lstlisting}[
    basicstyle=\ttfamily\scriptsize,
    frame=single,
    numbers=none,
    breaklines=true,
    showstringspaces=false
]
Pattern
1 USER:  INQUIRY
2 AGENT: ANSWER
3 USER:  DEFINITION REQUEST
4 AGENT: DEFINITION
\end{lstlisting}
\end{minipage}
\hfill
\begin{minipage}[t]{0.48\textwidth}
\captionsetup{type=figure}
\begin{lstlisting}[
    basicstyle=\ttfamily\scriptsize,
    frame=single,
    numbers=none,
    breaklines=true,
    showstringspaces=false
]
Example
1 USER: What did you need to talk about?
2 AGENT: I have some questions about my dossier.
3 USER: What's a dossier?
4 AGENT: A dossier is a collection of information about someone or something.
\end{lstlisting}
\end{minipage}
\caption{Definition Request.}
\label{lst:def_pattern}
\end{figure}

\begin{figure}[H]
\centering
\begin{minipage}[t]{0.48\textwidth}
\captionsetup{type=figure}
\begin{lstlisting}[
    basicstyle=\ttfamily\scriptsize,
    frame=single,
    numbers=none,
    breaklines=true,
    showstringspaces=false
]
Pattern
1 USER:  INQUIRY
2 AGENT: ANSWER
3 USER:  EXAMPLE REQUEST
4 AGENT: EXAMPLE
\end{lstlisting}
\end{minipage}
\hfill
\begin{minipage}[t]{0.48\textwidth}
\captionsetup{type=figure}
\begin{lstlisting}[
    basicstyle=\ttfamily\scriptsize,
    frame=single,
    numbers=none,
    breaklines=true,
    showstringspaces=false
]
Example
1 USER:  Have you been to Paris before?
2 AGENT: Many times. There are so many attractions worth visiting!
3 USER:  What are some examples?
4 AGENT: The Eiffel Tower, Louvre Museum, and Notre-Dame Cathedral are must-sees.
\end{lstlisting}
\end{minipage}
\caption{Example Request.}
\label{lst:example_pattern}
\end{figure}

\begin{figure}[H]
\centering
\begin{minipage}[t]{0.48\textwidth}
\captionsetup{type=figure}
\begin{lstlisting}[
    basicstyle=\ttfamily\scriptsize,
    frame=single,
    numbers=none,
    breaklines=true,
    showstringspaces=false
]
Pattern
1 USER:  INQUIRY
2 AGENT: RESPONSE
3 USER:  ACKNOWLEDGMENT/ASSESSMENT/LAUGHTER
4 AGENT: LAST TOPIC CHECK/NO RESPONSE
\end{lstlisting}
\end{minipage}
\hfill
\begin{minipage}[t]{0.48\textwidth}
\captionsetup{type=figure}
\begin{lstlisting}[
    basicstyle=\ttfamily\scriptsize,
    frame=single,
    numbers=none,
    breaklines=true,
    showstringspaces=false
]
Example
1 USER:  What is my out-of-pocket maximum?
2 AGENT: Your out-of-pocket maximum is $3000.
3 USER:  All right.
4 AGENT: Is there anything else you'd like to know?
\end{lstlisting}
\end{minipage}
\caption{Sequence Closer.}
\label{lst:closer_pattern}
\end{figure}

\begin{figure}[H]
\centering
\begin{minipage}[t]{0.48\textwidth}
\captionsetup{type=figure}
\begin{lstlisting}[
    basicstyle=\ttfamily\scriptsize,
    frame=single,
    numbers=none,
    breaklines=true,
    showstringspaces=false
]
Pattern
1 USER:  INQUIRY
2 AGENT: RESPONSE
3 USER:  ABORT
4 AGENT: ACKNOWLEDGMENT
\end{lstlisting}
\end{minipage}
\hfill
\begin{minipage}[t]{0.48\textwidth}
\captionsetup{type=figure}
\begin{lstlisting}[
    basicstyle=\ttfamily\scriptsize,
    frame=single,
    numbers=none,
    breaklines=true,
    showstringspaces=false
]
Example
1 USER:  Do you know anything about health food?
2 AGENT: I don't really eat health food.
3 USER:  Never mind.
4 AGENT: Is there anything else you'd like to know?
\end{lstlisting}
\end{minipage}
\caption{Sequence Abort.}
\label{lst:abort_pattern}
\end{figure}

\begin{figure}[H]
\centering
\begin{minipage}[t]{0.48\textwidth}
\captionsetup{type=figure}
\begin{lstlisting}[
    basicstyle=\ttfamily\scriptsize,
    frame=single,
    numbers=none,
    breaklines=true,
    showstringspaces=false
]
Pattern
1 USER:  INQUIRY (Grounded)
2 AGENT: ANSWER
\end{lstlisting}
\end{minipage}
\hfill
\begin{minipage}[t]{0.48\textwidth}
\captionsetup{type=figure}
\begin{lstlisting}[
    basicstyle=\ttfamily\scriptsize,
    frame=single,
    numbers=none,
    breaklines=true,
    showstringspaces=false
]
Example
1 USER:  On what movies is Spielberg an uncredited executive producer?
2 AGENT: Spielberg was an uncredited executive producer on The Haunting, The Prince of Egypt, Just Like Heaven, Shrek, Road to Perdition, and Evolution.
\end{lstlisting}
\end{minipage}
\caption{Inquiry (User) Grounded.}
\label{lst:inquiry_grounded_pattern}
\end{figure}

\begin{figure}[H]
\centering
\begin{minipage}[t]{0.48\textwidth}
\captionsetup{type=figure}
\begin{lstlisting}[
    basicstyle=\ttfamily\scriptsize,
    frame=single,
    numbers=none,
    breaklines=true,
    showstringspaces=false
]
Pattern
1 USER:  INQUIRY (Ungrounded)
2 AGENT: NONANSWER
\end{lstlisting}
\end{minipage}
\hfill
\begin{minipage}[t]{0.48\textwidth}
\captionsetup{type=figure}
\begin{lstlisting}[
    basicstyle=\ttfamily\scriptsize,
    frame=single,
    numbers=none,
    breaklines=true,
    showstringspaces=false
]
Example
1 USER:  How did German idealism influence the development of science in the 19th century?
2 AGENT: I don't have the information to answer that.
\end{lstlisting}
\end{minipage}
\caption{Inquiry (User) Ungrounded.}
\label{lst:inquiry_ungrounded_pattern}
\end{figure}

\begin{figure}[H]
\centering
\begin{minipage}[t]{0.48\textwidth}
\captionsetup{type=figure}
\begin{lstlisting}[
    basicstyle=\ttfamily\scriptsize,
    frame=single,
    numbers=none,
    breaklines=true,
    showstringspaces=false
]
Pattern
1 USER: PARTIAL REQUEST
2 AGENT: DETAIL REQUEST
\end{lstlisting}
\end{minipage}
\hfill
\begin{minipage}[t]{0.48\textwidth}
\captionsetup{type=figure}
\begin{lstlisting}[
    basicstyle=\ttfamily\scriptsize,
    frame=single,
    numbers=none,
    breaklines=true,
    showstringspaces=false
]
Example
1 USER: Can you recommend a smartphone?
2 AGENT: What brand do you want?
\end{lstlisting}
\end{minipage}
\caption{Detail Request All (Same pattern as Detail Request Partial).}
\label{lst:inquiry_detail_request_all_pattern}
\end{figure}

\begin{figure}[H]
\centering
\begin{minipage}[t]{0.48\textwidth}
\captionsetup{type=figure}
\begin{lstlisting}[
    basicstyle=\ttfamily\scriptsize,
    frame=single,
    numbers=none,
    breaklines=true,
    showstringspaces=false
]
Pattern
1 USER: PARTIAL REQUEST
2 AGENT: DETAIL REQUEST
3 USER: DETAIL
4 AGENT: DETAIL REQUEST
\end{lstlisting}
\end{minipage}
\hfill
\begin{minipage}[t]{0.48\textwidth}
\captionsetup{type=figure}
\begin{lstlisting}[
    basicstyle=\ttfamily\scriptsize,
    frame=single,
    numbers=none,
    breaklines=true,
    showstringspaces=false
]
Example
1 USER:  We want to do a granite countertop.
2 AGENT: Are you gonna want a backsplash too?
3 USER:  Yes, definitely!
4 AGENT: Over- or undermounted sink?
\end{lstlisting}
\end{minipage}
\caption{Detail Request Expanded.}
\label{lst:inquiry_detail_request_expanded_pattern}
\end{figure}

\begin{figure}[H]
\centering
\begin{minipage}[t]{0.48\textwidth}
\captionsetup{type=figure}
\begin{lstlisting}[
    basicstyle=\ttfamily\scriptsize,
    frame=single,
    numbers=none,
    breaklines=true,
    showstringspaces=false
]
Pattern
1 USER:  PARTIAL REQUEST
2 AGENT: DETAIL REQUEST
\end{lstlisting}
\end{minipage}
\hfill
\begin{minipage}[t]{0.48\textwidth}
\captionsetup{type=figure}
\begin{lstlisting}[
    basicstyle=\ttfamily\scriptsize,
    frame=single,
    numbers=none,
    breaklines=true,
    showstringspaces=false
]
Example
1 USER:  How many days do I get for vacation if 
         I've worked here for 3 years?
2 AGENT: What country do you work in?

\end{lstlisting}
\end{minipage}
\caption{Detail Request Partial (Same pattern as Detail Request All).}
\label{lst:inquiry_detail_request_partial_pattern}
\end{figure}

\begin{figure}[H]
\centering
\begin{minipage}[t]{0.48\textwidth}
\captionsetup{type=figure}
\begin{lstlisting}[
    basicstyle=\ttfamily\scriptsize,
    frame=single,
    numbers=none,
    breaklines=true,
    showstringspaces=false
]
Pattern
1 USER:  PARTIAL REQUEST
2 AGENT: DETAIL REQUEST
3 USER:  DETAIL REQUEST
4 AGENT: DETAIL
\end{lstlisting}
\end{minipage}
\hfill
\begin{minipage}[t]{0.48\textwidth}
\captionsetup{type=figure}
\begin{lstlisting}[
    basicstyle=\ttfamily\scriptsize,
    frame=single,
    numbers=none,
    breaklines=true,
    showstringspaces=false
]
Example
1 USER:  I need to get life insurance with my  
         children as beneficiaries.
2 AGENT: What type of policy?
3 USER:  What are my choices?
4 AGENT: Term or whole life.
\end{lstlisting}
\end{minipage}
\caption{Expansion Choices.}
\label{lst:inquiry_expansion_choices_pattern}
\end{figure}

\begin{figure}[H]
\centering
\begin{minipage}[t]{0.48\textwidth}
\captionsetup{type=figure}
\begin{lstlisting}[
    basicstyle=\ttfamily\scriptsize,
    frame=single,
    numbers=none,
    breaklines=true,
    showstringspaces=false
]
Pattern
1 USER:  PARTIAL REQUEST
2 AGENT: DETAIL REQUEST
3 USER:  REPAIR INITIATOR
4 AGENT: REPAIR
\end{lstlisting}
\end{minipage}
\hfill
\begin{minipage}[t]{0.48\textwidth}
\captionsetup{type=figure}
\begin{lstlisting}[
    basicstyle=\ttfamily\scriptsize,
    frame=single,
    numbers=none,
    breaklines=true,
    showstringspaces=false
]
Example
1 USER:  I would like to book a trip to South 
         America!
2 AGENT: Okay, what kinds of things do you like 
         to do on vacation?
3 USER:  Can you give an example?
4 AGENT: Hiking, museums, nightlife, etc.?

\end{lstlisting}
\end{minipage}
\caption{Expansion Repair.}
\label{lst:inquiry_expansion_repair_pattern}
\end{figure}

\begin{figure}[H]
\centering
\begin{minipage}[t]{0.48\textwidth}
\captionsetup{type=figure}
\begin{lstlisting}[
    basicstyle=\ttfamily\scriptsize,
    frame=single,
    numbers=none,
    breaklines=true,
    showstringspaces=false
]
Pattern
1 USER:  DETAIL
2 AGENT: CONTINUER
\end{lstlisting}
\end{minipage}
\hfill
\begin{minipage}[t]{0.48\textwidth}
\captionsetup{type=figure}
\begin{lstlisting}[
    basicstyle=\ttfamily\scriptsize,
    frame=single,
    numbers=none,
    breaklines=true,
    showstringspaces=false
]
Example
1 USER:  Oh great! You do pour over coffee!
2 AGENT: Yes, we do!

\end{lstlisting}
\end{minipage}
\caption{Preliminary Detail.}
\label{lst:inquiry_preliminary_detail_pattern}
\end{figure}

\begin{figure}[H]
\centering
\begin{minipage}[t]{0.48\textwidth}
\captionsetup{type=figure}
\begin{lstlisting}[
    basicstyle=\ttfamily\scriptsize,
    frame=single,
    numbers=none,
    breaklines=true,
    showstringspaces=false
]
Pattern
1 USER:  PRE-REQUEST
2 AGENT: GO AHEAD
\end{lstlisting}
\end{minipage}
\hfill
\begin{minipage}[t]{0.48\textwidth}
\captionsetup{type=figure}
\begin{lstlisting}[
    basicstyle=\ttfamily\scriptsize,
    frame=single,
    numbers=none,
    breaklines=true,
    showstringspaces=false
]
Example
1 USER:  Is this the IT help desk?
2 AGENT: Yes, it is!
\end{lstlisting}
\end{minipage}
\caption{Preliminary Screen.}
\label{lst:inquiry_preliminary_screen_pattern}
\end{figure}

\begin{figure}[H]
\centering
\begin{minipage}[t]{0.48\textwidth}
\captionsetup{type=figure}
\begin{lstlisting}[
    basicstyle=\ttfamily\scriptsize,
    frame=single,
    numbers=none,
    breaklines=true,
    showstringspaces=false
]
Pattern
1 USER:  FULL REQUEST
2 AGENT: GRANT
\end{lstlisting}
\end{minipage}
\hfill
\begin{minipage}[t]{0.48\textwidth}
\captionsetup{type=figure}
\begin{lstlisting}[
    basicstyle=\ttfamily\scriptsize,
    frame=single,
    numbers=none,
    breaklines=true,
    showstringspaces=false
]
Example
1 USER:  We are looking for a single family home 
         with three bedrooms and that's a good 
         school district.
2 AGENT: Okay, I have a few houses you might be 
         interested in.

\end{lstlisting}
\end{minipage}
\caption{Recommendation Compact.}
\label{lst:recommendation_compact_pattern}
\end{figure}

\begin{figure}[H]
\centering
\begin{minipage}[t]{0.48\textwidth}
\captionsetup{type=figure}
\begin{lstlisting}[
    basicstyle=\ttfamily\scriptsize,
    frame=single,
    numbers=none,
    breaklines=true,
    showstringspaces=false
]
Pattern
1 USER:  PARTIAL REQUEST
2 AGENT: DETAIL REQUEST
3 USER:  DETAIL
4 AGENT: DETAIL REQUEST
5 USER:  DETAIL
6 AGENT: GRANT
\end{lstlisting}
\end{minipage}
\hfill
\begin{minipage}[t]{0.48\textwidth}
\captionsetup{type=figure}
\begin{lstlisting}[
    basicstyle=\ttfamily\scriptsize,
    frame=single,
    numbers=none,
    breaklines=true,
    showstringspaces=false
]
Example
1 USER:  Give me a drip coffee please.
2 AGENT: Would you like cream or sugar?
3 USER:  Whole milk please, no sugar.
4 AGENT: What size?
5 USER:  Large.
6 AGENT: Okay, that will be $3.65 please.

\end{lstlisting}
\end{minipage}
\caption{Recommendation Expanded.}
\label{lst:recommendation_expanded_pattern}
\end{figure}

\begin{figure}[H]
\centering
\begin{minipage}[t]{0.48\textwidth}
\captionsetup{type=figure}
\begin{lstlisting}[
    basicstyle=\ttfamily\scriptsize,
    frame=single,
    numbers=none,
    breaklines=true,
    showstringspaces=false
]
Pattern
1 USER:  FULL REQUEST
2 AGENT: GRANT
3 USER:  INCREMENTAL REQ.
4 AGENT: GRANT

\end{lstlisting}
\end{minipage}
\hfill
\begin{minipage}[t]{0.48\textwidth}
\captionsetup{type=figure}
\begin{lstlisting}[
    basicstyle=\ttfamily\scriptsize,
    frame=single,
    numbers=none,
    breaklines=true,
    showstringspaces=false
]
Example
1 USER:  Can you recommend a Mexican restaurant   
         within walking distance?
2 AGENT: Mario's in just two blocks down Beach  
         Blvd.
3 USER:  What about pizza?
4 AGENT: Munchies Pizza is one block west on 
         Main Street.

\end{lstlisting}
\end{minipage}
\caption{Recommendation Incremental.}
\label{lst:recommendation_incremental_pattern}
\end{figure}

\clearpage
\section{Judge Validation}
\label{sec:judge_validation}

To assess the reliability of the LLM-as-a-judge pipeline, we conducted a human validation study on a stratified sample of 60 model responses drawn from the full benchmark.

\paragraph{Sampling Strategy.}
We sampled exactly 20 responses per set (Basic, RAG, Complex Request) for a total of 60. Within each set, samples are distributed evenly across all patterns: patterns with higher failure rates (\textit{Repeat Request}, \textit{Sequence Closer}, \textit{Sequence Abort}, \textit{Inquiry (Ungrounded)}) receive 3 samples each, while remaining patterns receive 1 to 2 samples. Samples are drawn from at least two different model families per pattern to avoid model-specific bias. Table~\ref{tab:annotation_sample} summarizes the distribution.

\begin{table}[H]
\centering
\caption{Distribution of the 60-sample human validation set (20 per set).}
\label{tab:annotation_sample}
\small
\begin{tabular}{lr}
\toprule
\textbf{Pattern} & \textbf{n} \\
\midrule
Repeat Request       & 8 \\
Sequence Closer      & 8 \\
Sequence Abort       & 8 \\
Inquiry / Incremental& 5 \\
Paraphrase Request   & 5 \\
Definition Request   & 5 \\
Example Request      & 5 \\
Recommendation       & 3 \\
Detail Request       & 3 \\
Self-Correction      & 3 \\
Inquiry (Ungrounded) & 3 \\
Expansion            & 2 \\
Preliminary          & 2 \\
\midrule
\textbf{Total}       & \textbf{60} \\
\bottomrule
\end{tabular}
\end{table}

\paragraph{Annotation Procedure.}
Each sample presents a conversation context and the model's generated response. A human annotator independently assigned one label from a small, pattern-specific set of conversational action categories (e.g., for \textit{Repeat Request}: \textit{Repeat}, \textit{Paraphrase}, \textit{Answer}, \textit{Other}), mirroring the classification schema of the judge LLM, without consulting judge labels beforehand. The same 60 samples were then classified by an independent Llama-3.3-70B judge using the same prompt.

\paragraph{Agreement Results.}
We evaluate agreement at two levels. First, we compute Cohen's $\kappa$ on the raw predicted label (Table~\ref{tab:judge_agreement}, left). Second, and more directly relevant to benchmark validity, we evaluate agreement on the \emph{pass/fail} outcome: whether the predicted label falls within the set of acceptable conversational actions for each pattern, as defined by our aggregator (Table~\ref{tab:judge_agreement}, right). This second metric better reflects what actually matters for scoring --- two raters may assign different fine-grained labels yet agree that the response is or is not acceptable.

At the label level, both judges achieve substantial agreement with the human annotator ($\kappa = 0.749$ for Mistral-Large; $\kappa = 0.824$ for Llama-3.3-70B). Inter-judge agreement is high ($\kappa = 0.806$, 83.3\%). At the pass/fail level, agreement improves markedly: Human--Mistral reaches $\kappa = 0.899$ (95.0\%), Human--Llama reaches $\kappa = 0.897$ (95.0\%), and inter-judge agreement reaches $\kappa = 0.863$ (93.3\%). This confirms that the label-level disagreements are largely cases where different valid labels map to the same scoring outcome, rather than genuine errors in the judge pipeline.

\begin{table}[H]
\centering
\caption{Cohen's $\kappa$ between rater pairs at label and pass/fail level.}
\label{tab:judge_agreement}
\small
\setlength{\tabcolsep}{4pt}
\begin{tabular}{lcccc}
\toprule
 & \multicolumn{2}{c}{\textbf{Label}} & \multicolumn{2}{c}{\textbf{Pass/fail}} \\
\cmidrule(lr){2-3}\cmidrule(lr){4-5}
\textbf{Pair} & $\kappa$ & \% & $\kappa$ & \% \\
\midrule
H vs.\ Mistral   & 0.749 & 78.3 & 0.899 & 95.0 \\
H vs.\ Llama     & 0.824 & 85.0 & 0.897 & 95.0 \\
Mistral vs.\ Llama & 0.806 & 83.3 & 0.863 & 93.3 \\
\bottomrule
\end{tabular}
\end{table}

\end{document}